\documentclass{article}
\pdfoutput=1
\usepackage[nonatbib,preprint]{neurips_2023}

\usepackage[utf8]{inputenc} 
\usepackage[T1]{fontenc}    
\usepackage{hyperref}       
\usepackage{url}            
\usepackage{booktabs}       
\usepackage{amsfonts}       
\usepackage{nicefrac}       
\usepackage{microtype}      
\usepackage{xcolor,colortbl}         

\usepackage{tabularx}
\usepackage{adjustbox}
\usepackage{xspace}
\usepackage{enumitem}
\usepackage{amssymb}
\usepackage{amsmath}
\usepackage{multirow}
\usepackage{graphicx}
\usepackage[center]{subfigure}
\usepackage{wrapfig}
\usepackage[]{algorithmic} 
\usepackage{tablefootnote}
\usepackage{threeparttable}
\usepackage{makecell}
\usepackage[font={small}]{caption}
\usepackage{diagbox}
\usepackage{empheq}
\usepackage{amsthm}
\usepackage{todonotes}
\usepackage{algorithm}

\title{Multimodal Graph Learning for Generative Tasks}

\author{%
  Minji Yoon \\
  Carnegie Mellon University\\
  \And
  Jing Yu Koh \\
  Carnegie Mellon University\\
  \And
  Bryan Hooi\\
  National University of Singapore\\
  \And
  Ruslan Salakhutdinov\\
  Carnegie Mellon University\\
}

\begin{document}
\maketitle

\newcommand{\method}{\textsc{KTN}\xspace}
\renewcommand{\qedsymbol}{$\blacksquare$}

\newtheorem{problem_inf}{Informal Problem Definition}
\newtheorem{problem}{Problem Definition}
\newtheorem{theorem}{Theorem}
\newtheorem{obs}{Observation}

\newcommand{\john}[1]{\todo[inline,color=green!20!white]{\textbf{John:} #1}}
\newcommand{\bryan}[1]{\todo[inline,color=blue!20!white]{\textbf{Bryan:} #1}}
\newcommand{\dustin}[1]{\todo[inline,color=red!20!white]{\textbf{Dustin:} #1}}
\newcommand{\minji}[1]{\textcolor{blue}{#1}}

\newcommand{\concat}{\mathbin\Vert}

\newcommand{\snre}{\sigma_e}
\newcommand{\snrf}{\sigma_f}

\begin{abstract}
Multimodal learning combines multiple data modalities, broadening the types and complexity of data our models can utilize: for example, from plain text to image-caption pairs.
Most multimodal learning algorithms focus on modeling simple one-to-one pairs of data from two modalities, such as image-caption pairs, or audio-text pairs.
However, in most real-world settings, entities of different modalities interact with each other in more complex and multifaceted ways, going beyond one-to-one mappings. 
We propose to represent these complex relationships as graphs, allowing us to capture data with any number of modalities, and with complex relationships between modalities that can flexibly vary from one sample to another.
Toward this goal, we propose Multimodal Graph Learning (MMGL), a general and systematic framework for capturing information from multiple multimodal neighbors with relational structures among them.
In particular, we focus on MMGL for \emph{generative} tasks, building upon pretrained Language Models (LMs), aiming to augment their text generation with multimodal neighbor contexts.
We study three research questions raised by MMGL: 
(1) how can we infuse multiple neighbor information into the pretrained LMs, while avoiding scalability issues?
(2) how can we infuse the graph structure information among multimodal neighbors into the LMs?
and (3) how can we finetune the pretrained LMs to learn from the neighbor context in a parameter-efficient manner?
We conduct extensive experiments to answer these three questions on MMGL and analyze the empirical results to pave the way for future MMGL research.
\end{abstract}

\section{Introduction}
\label{sec:introduction}
There are diverse data modalities in real-world applications, from commonly observed texts, images, and videos to time series data or domain-specific modalities like protein sequences.
These various modalities are not collected individually but together with multifaceted relations among them.
Wikipedia~\cite{burns2023wiki} is one of the most popular sources of multimodal web content, providing multimodal data such as texts, images, and captions.
TimeBuilder~\cite{tan2023timelineqa}, recently released by Meta, builds personal timelines using each user's multimodal data, including their photos, maps, shopping, and music history.
In addition to these examples, important industrial and medical decisions are also made by considering diverse multimodal data such as images, tables, or audio~\cite{huang2020review,singh2007multimodal}.
These multimodal data have complicated $many$-to-$many$ relations among their multimodal entities --- which can be represented as graphs --- providing open research space on how to understand them holistically.

\begin{figure*}[t!]
    \centering
    \subfigure[$1$-to-$1$ Multimodal Learning]
    {
    \label{fig:data:1to1}
    \includegraphics[width=.27\linewidth]{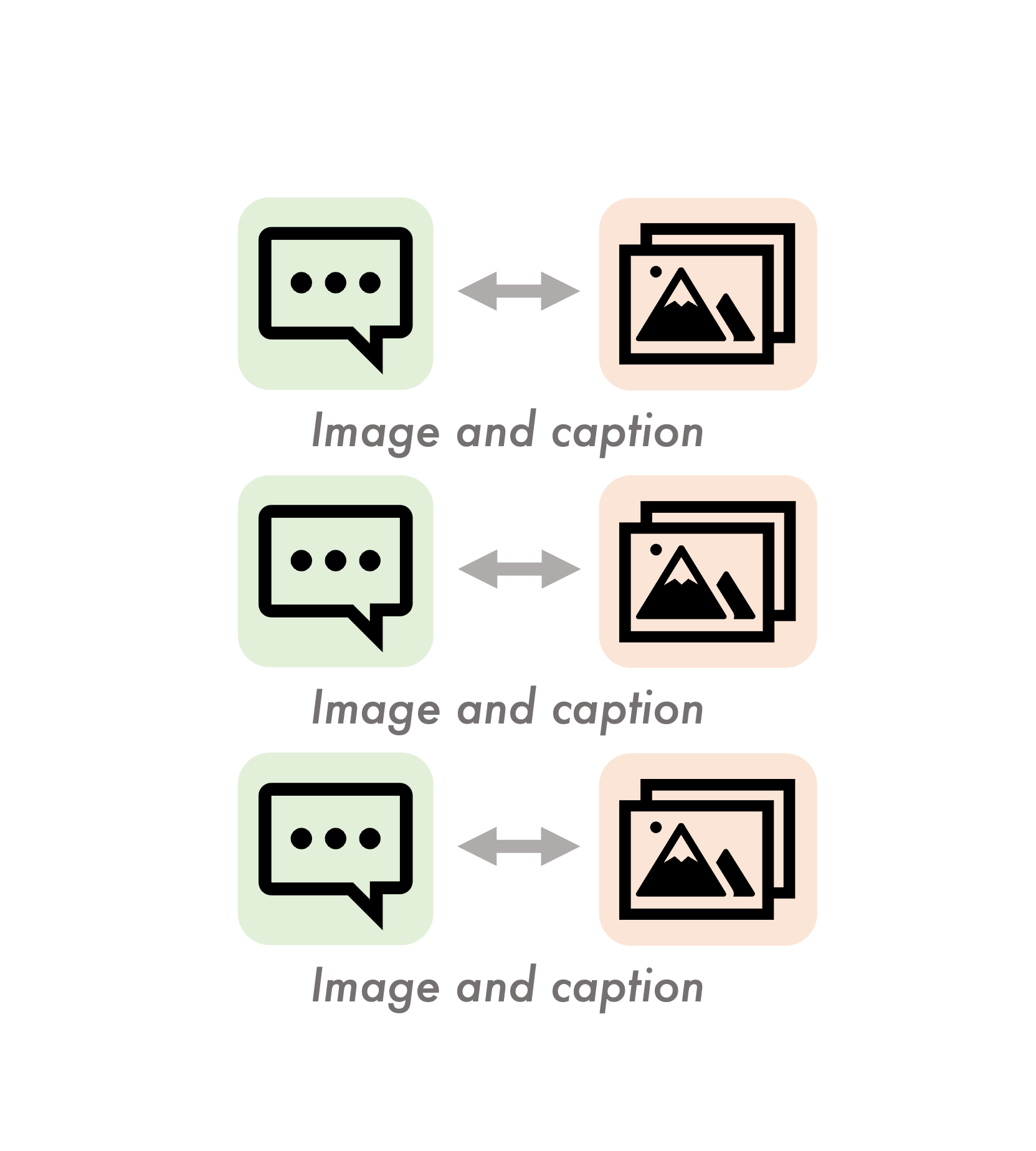}
    }
    \subfigure[$many$-to-$many$ Multimodal Graph Learning]
    {
    \label{fig:data:nton}
    \includegraphics[width=.69\linewidth]{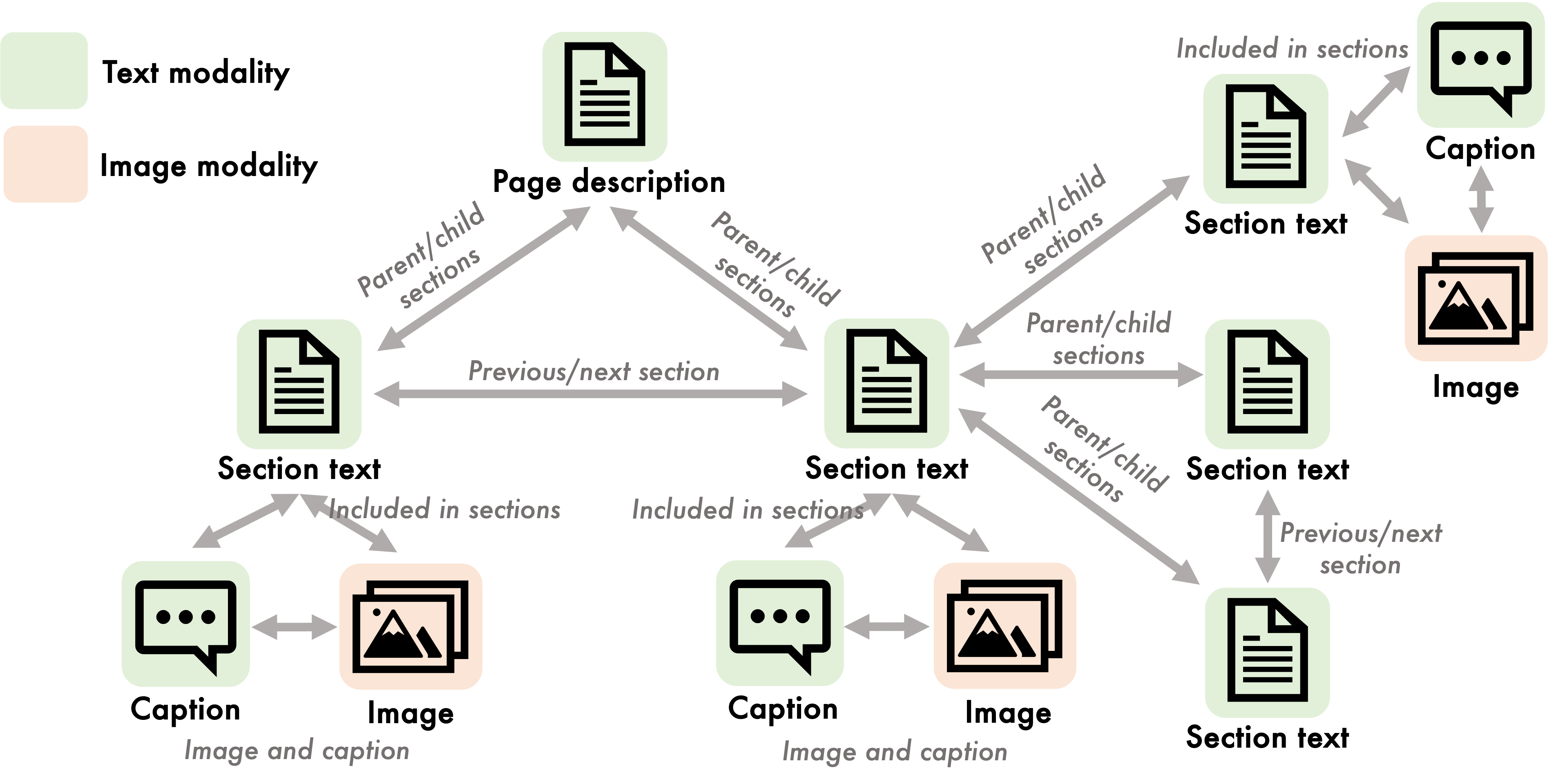}
    }
    \caption
    { \small
    \textbf{Multimodal datasets extracted from Wikipedia}:
    (a) Most multimodal models target multimodal datasets with clear $1$-to-$1$ mappings between modalities.
    (b) Multimodal Graph Learning (MMGL) handles multimodal datasets with complicated relations among multiple multimodal neighbors.
    }
    \label{fig:data}
 \end{figure*}

With the rise of multimodal datasets, various ground-breaking research has been done in multimodal learning.
Previously, multimodal learning focused on novel architectures, extending transformers~\cite{hendricks2021decoupling, liang2022high,tsai2019multimodal} or graph neural networks~\cite{hu2020heterogeneous,schlichtkrull2018modeling}, and training them from scratch using large-scaled multimodal datasets.
Fueled by the strong generative power of pretrained Language Models (LMs), recent multimodal approaches \cite{alayrac2022flamingo,li2023blip, koh2023generating} are built upon pretrained LMs and focus on the generation of multimodal content.
For instance, \cite{koh2023generating} generates images/text grounded on given text/images using pretrained image encoders and LMs.
However, all existing models assume that a pair of modalities with a clear $1$-to-$1$ mapping is provided as input (e.g., image-caption pairs in Figure~\ref{fig:data:1to1}). 
As a result, they cannot be directly applied on multimodal datasets with more general $many$-to-$many$ mappings among modalities (e.g., multimodal Wikipedia webpage in Figure~\ref{fig:data:nton}).

Here, we expand the scope of multimodal learning beyond $1$-to-$1$ mappings into multimodal graph learning (MMGL) while preserving generative abilities by integrating them into pretrained LMs.
We introduce a systematic framework on how MMGL processes multimodal neighbor information with graph structures among them and generate free-form texts using pretrained LMs (Figure~\ref{fig:flow}).
Our MMGL framework extracts \emph{neighbor encodings}, combines them with \emph{graph structure information}, and optimizes the model using \emph{parameter-efficient fine-tuning}.
Accordingly, we define three design spaces to study three research questions for MMGL as follows:
\begin{itemize}[leftmargin=7pt,topsep=0pt,itemsep=-1ex,partopsep=1ex,parsep=1ex]
    \item {
    \textbf{Research Question 1.}
    How can we provide multiple multimodal neighbor information to LMs while avoiding scalability issues? 
    }
    \item{
    \textbf{Research Question 2.}
    How can we infuse graph structure information among multimodal neighbors into LMs?
    }
    \item{
    \textbf{Research Question 3.}
    How can we finetune pretrained LMs to learn through multimodal neighbor information in parameter-efficient ways?
    }
\end{itemize}
In conventional multimodal learning with the $1$-to-$1$ mapping assumption, typically only one neighbor is provided (e.g., an image for a text caption)~\cite{alayrac2022flamingo, koh2023generating,li2023blip}. 
On the contrary, MMGL requires the processing of several neighbors with various data sizes (e.g., image resolution and text sequences of various lengths), which leads to the scalability issue.
For \textit{Research Question 1}, we study three neighbor encoding models: 
(1) \textit{Self-Attention with Text + Embeddings} (SA-Text+Embeddings) precomputes image embeddings using frozen encoders, then concatenates them to the input text sequences with any raw text from neighbors (originally proposed from~\cite{tsimpoukelli2021multimodal}), 
(2) \textit{Self-Attention with Embeddings} (SA-Embeddings) precomputes embeddings for both text and image modalities using frozen encoders and concatenates to the input text, 
and (3) \textit{Cross-Attention with Embeddings} (CA-Embeddings) feeds precomputed text or image embeddings into cross-attention layers of LMs.

In \textit{Research Question 2}, we study how to infuse graph structure information among multimodal neighbors into LMs (e.g., section hierarchy and image orders in Figure~\ref{fig:data:nton}).
We compare the sequential position encoding with two graph position encodings widely used in graph transformers~\cite{rampavsek2022recipe,ying2021transformers}:
\textit{Laplacian eigenvector position encoding} (LPE)~\cite{dwivedi2020generalization} and \textit{graph neural networks encoding} (GNN)~\cite{kipf2016semi} that runs GNNs on precomputed neighbor embeddings using graphs structures before feeding them into LMs.

\textit{Research Question 3} seeks to improve the cost and memory efficiency compared to full fine-tuning of LMs.
In this work, we explore three parameter-efficient fine-tuning (PEFT) methods~\cite{houlsby2019parameter}: \textit{Prefix tuning}~\cite{li2021prefix}, \textit{LoRA}~\cite{hu2021lora}, and \textit{Flamingo tuning}~\cite{alayrac2022flamingo}.
Which PEFT methods to use depends on the neighbor encoding model:
when neighbor information is concatenated into the input sequences (\textit{SA-Text+Embeddings} or \textit{SA-Embeddings} neighbor encodings), we can apply \textit{Prefix tuning} or \textit{LoRA} for fine-tuning.
When neighbor information is fed into cross-attention layers (\textit{CA-Embeddings} neighbor encoding), we apply \textit{Flamingo tuning} that finetunes only cross-attention layers with gating modules for stable finetuning~\cite{alayrac2022flamingo}.

Based on our MMGL framework, we run extensive experiments on the recently released multimodal dataset, WikiWeb2M~\cite{burns2023wiki}.
WikiWeb2M uniﬁes each Wikipedia webpage content to include all text, images, and their structures in a single example.
This makes it useful for studying multimodal content understanding with many-to-many text and image relationships, in the context of generative tasks.
Here, we focus on the section summarization task that aims to generate a sentence that captures information about the contents of one section by understanding the multimodal content on each Wikipedia page.
Through rigorous testing on WikiWeb2M, we provide intuitive empirical answers to research questions raised in MMGL. 

In summary, our contributions are:
\begin{itemize}[leftmargin=7pt,topsep=0pt,itemsep=-1ex,partopsep=1ex,parsep=1ex]
    \item {
    \textbf{Multimodal Graph Learning (MMGL):} 
    We introduce a systematic MMGL framework for processing multimodal neighbor information with graph structures among them, and generating free-form texts using pretrained LMs.
    }
    \item{
    \textbf{Principled Research Questions:}
    We introduce three research problems MMGL is required to answer: (1) how to provide multiple neighbor information to the pretrained LMs, (2) how to infuse graph structure information into LMs, and (3) how to fine-tune the LMs parameter-efficiently. 
    This paves research directions for future MMGL research.
    }
    \item{
    \textbf{Extensive Empirical Results:} 
    We show empirically that (1) neighbor context improves generation performance, (2) \textit{SA-Text+Embeddings} neighbor encoding shows the highest performance while sacrificing the scalability, (3) \textit{GNN} embeddings are the most effective graph position encodings, and (4) \textit{SA-Text+Embeddings} neighbor encoding with \textit{LoRA} and \textit{CA-Embeddings} neighbor encoding with \textit{Flamingo tuning} show the highest performance among different PEFT models.
    }
\end{itemize}
Our code is publicly available at~\footnote{~\url{https://github.com/minjiyoon/MMGL}}.

\section{Related Work}
\label{sec:related_work}
\paragraph{End-to-End Multimodal Learning:}
While many discriminative multimodal models~\cite{jia2021scaling,radford2021learning} have also been developed, we primarily consider related work on generative multimodal models, as this is most closely related with our approach. Several recent approaches tackle multimodal learning by building upon the Transformer~\cite{vaswani2017attention} architecture. Multimodal extensions typically use either full self-attention over modalities
concatenated across the sequence dimension~\cite{chen2020uniter,su2019vl} or a cross-modal attention layer~\cite{tsai2019multimodal}.
Self-supervised multimodal pretraining methods train these architectures from large-scale unlabeled multimodal data before transferring them to downstream multimodal tasks via fine-tuning~\cite{hendricks2021decoupling, liang2022high}.
These methods perform end-to-end pre-training, incurring extremely high computation costs, especially as model parameters increase~\cite{li2023blip}.
Moreover, this framework is relatively inflexible for end-to-end pre-trained models to leverage readily available unimodal pre-trained models, such as text-only LMs or pretrained vision models.

\paragraph{Multimodal Learning with Frozen Image Encoders and Large Language Models:}
Recently, various vision-language models have been proposed to leverage off-the-shelf pre-trained models and keep them frozen during pretrainig~\cite{alayrac2022flamingo,li2023blip, koh2023generating}.
To input visual information directly to a frozen text-only LLM, a key challenge is to align visual features to the text space.
Motivated by Frozen~\cite{tsimpoukelli2021multimodal}, which finetunes a visual encoder to map images into the hidden space of a text-only LLM, Blip-2~\cite{li2023blip} and GILL~\cite{koh2023generating} finetune separate image mapping networks whose inputs are precomputed by frozen image encoders and outputs are directly used as soft prompts to LLMs.
On the other hand, Flamingo~\cite{alayrac2022flamingo} inserts new cross-attention layers into the LLM to inject visual features and pre-trains the new layers on image-text pairs.
Note that all these methods primarily focus on processing \textit{interleaved image and text inputs} to generate text outputs. 

\paragraph{Graph Neural Networks on Multimodal Graphs}
Heterogeneous Graph Neural Networks (HGNNs) extend Graph Neural Networks (GNNs)~\cite{zhang2019graph} to learn from multimodal heterogeneous graphs. This is done through precomputing input node embeddings using frozen encoders, and training the GNN to map different modality embeddings either at the input layer~\cite{schlichtkrull2018modeling}, intermediate~\cite{hu2020heterogeneous}, or late layers~\cite{yoon2022zero}.
However, most HGNN models focus on node classification, and are difficult to adapt for generative tasks.
Recently, various approaches have been proposed to fine-tune LLMs with GNNs on text-attributed graphs~\cite{chien2021node, he2023explanations, zhao2022learning}.
These methods specialize in node/edge classification tasks by putting GNN models after LLMs, making them difficult to adapt for use in generative tasks.

\section{Multimodal Graph Learning for Generative Tasks}
\label{sec:proposed_work}
\begin{figure*}[t!]
    \centering
    \subfigure[Multimodal neighbors with graph structures]
    {
    \label{fig:flow:input}
    \includegraphics[width=.28\linewidth]{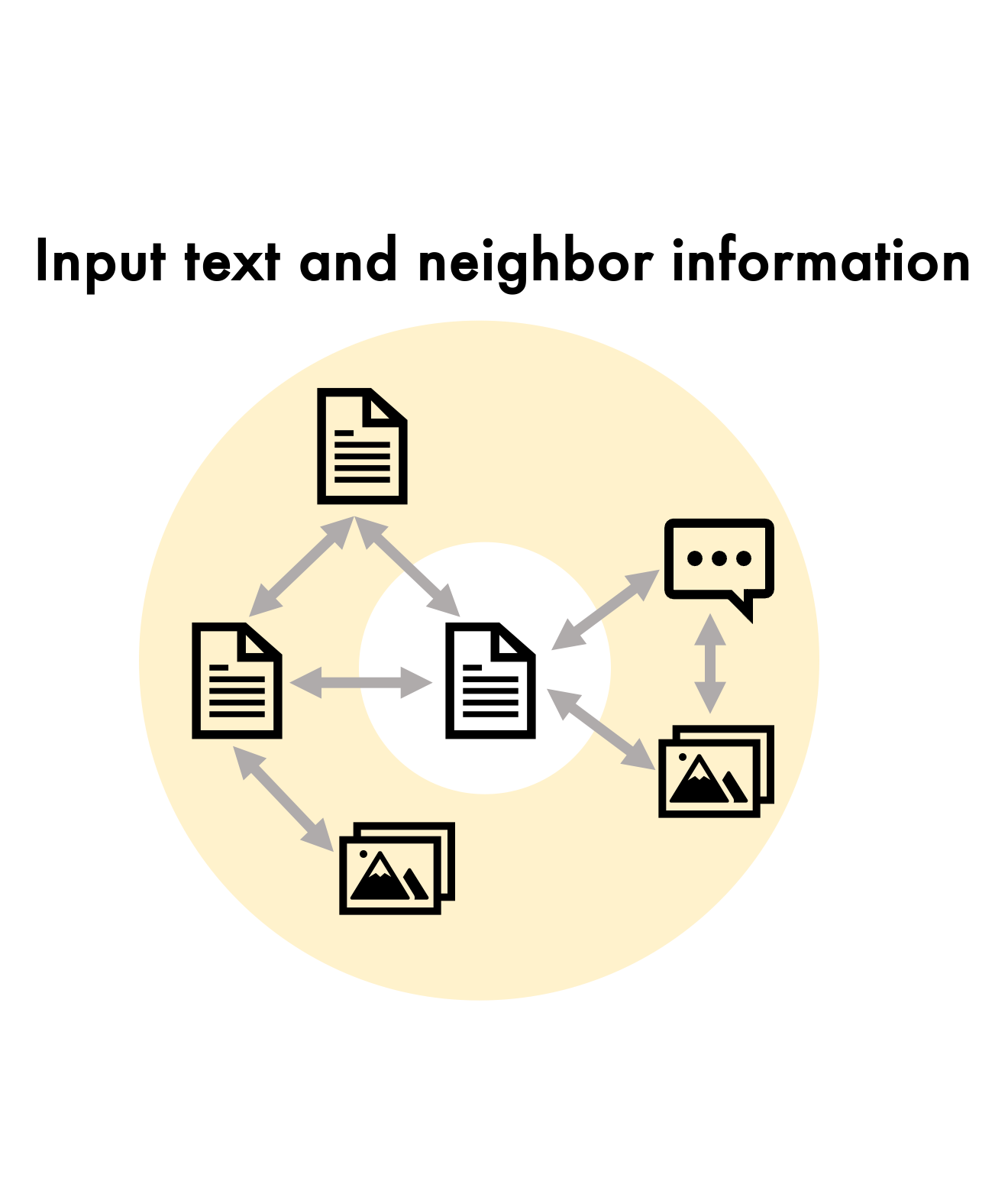}
    }
    \hspace{5mm}
    \subfigure[Multimodal neighbor encoding using frozen vision/text encoders]
    {
    \label{fig:flow:encode}
    \includegraphics[width=.6\linewidth]{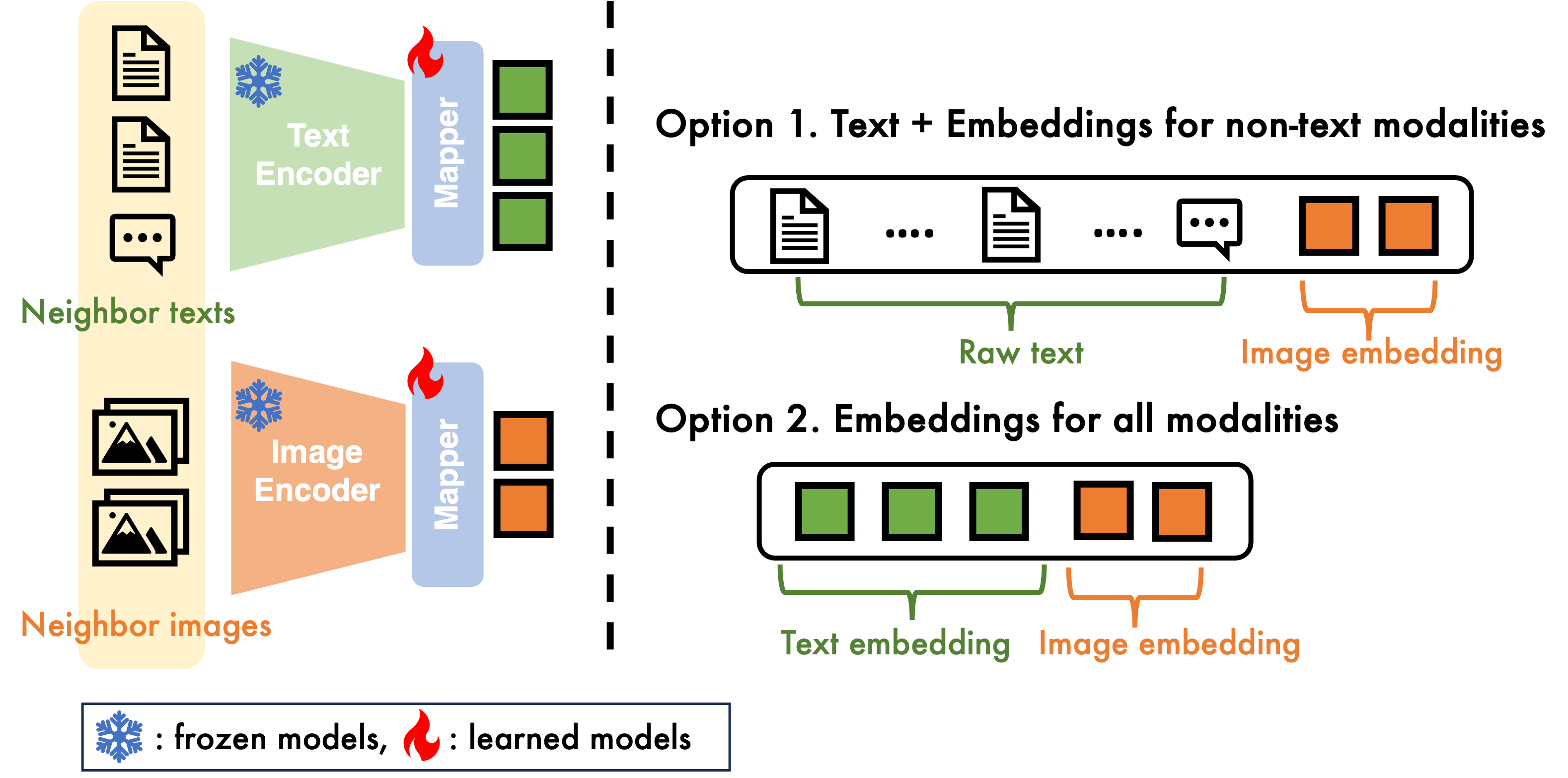}
    }
    \subfigure[Graph position encodings]
    {
    \label{fig:flow:graph}
    \includegraphics[width=.21\linewidth]{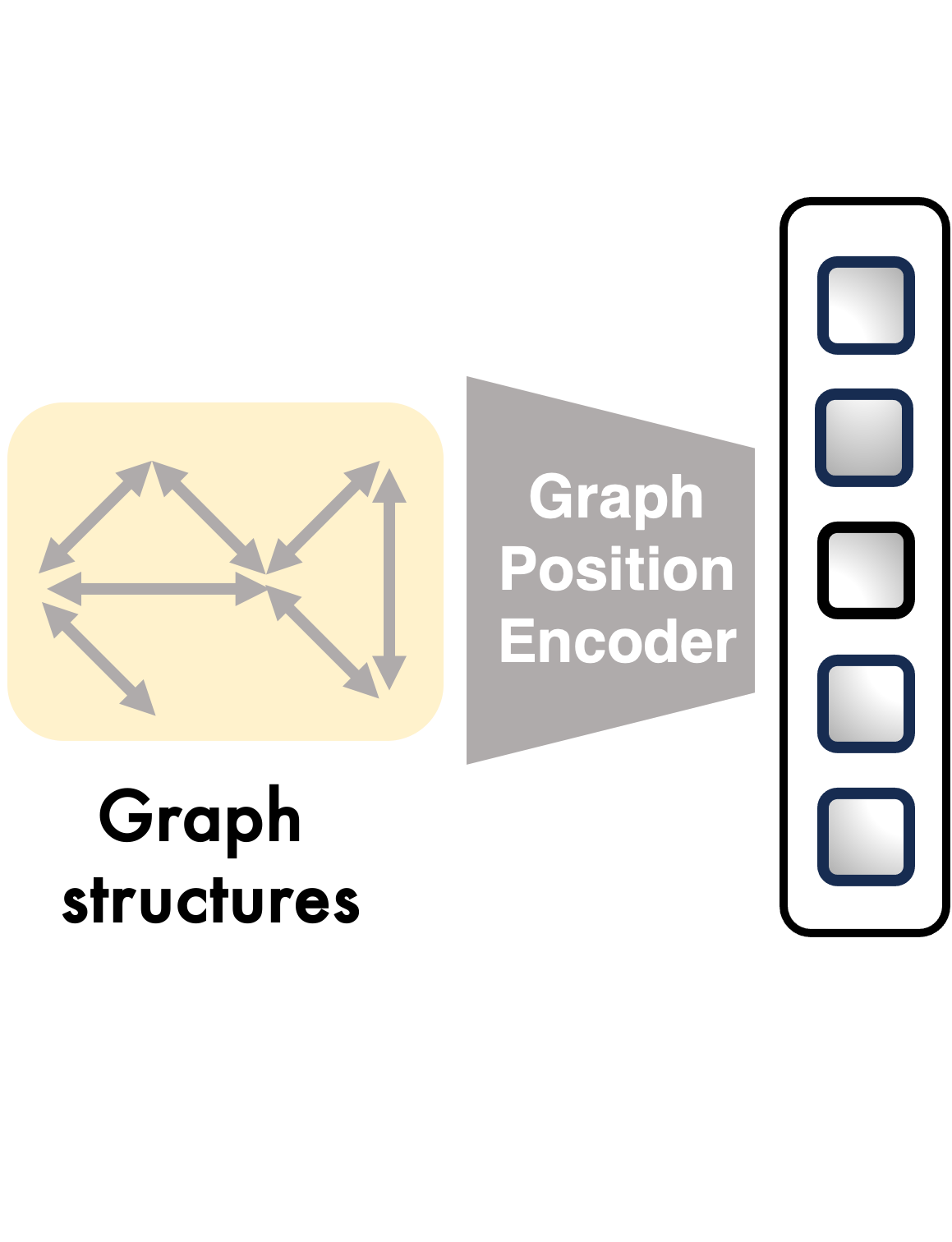}
    }
    \hspace{2mm}
    \subfigure[Text generation using frozen/finetuned LMs and encoded neighbor information]
    {
    \label{fig:flow:decode}
    \includegraphics[width=.73\linewidth]{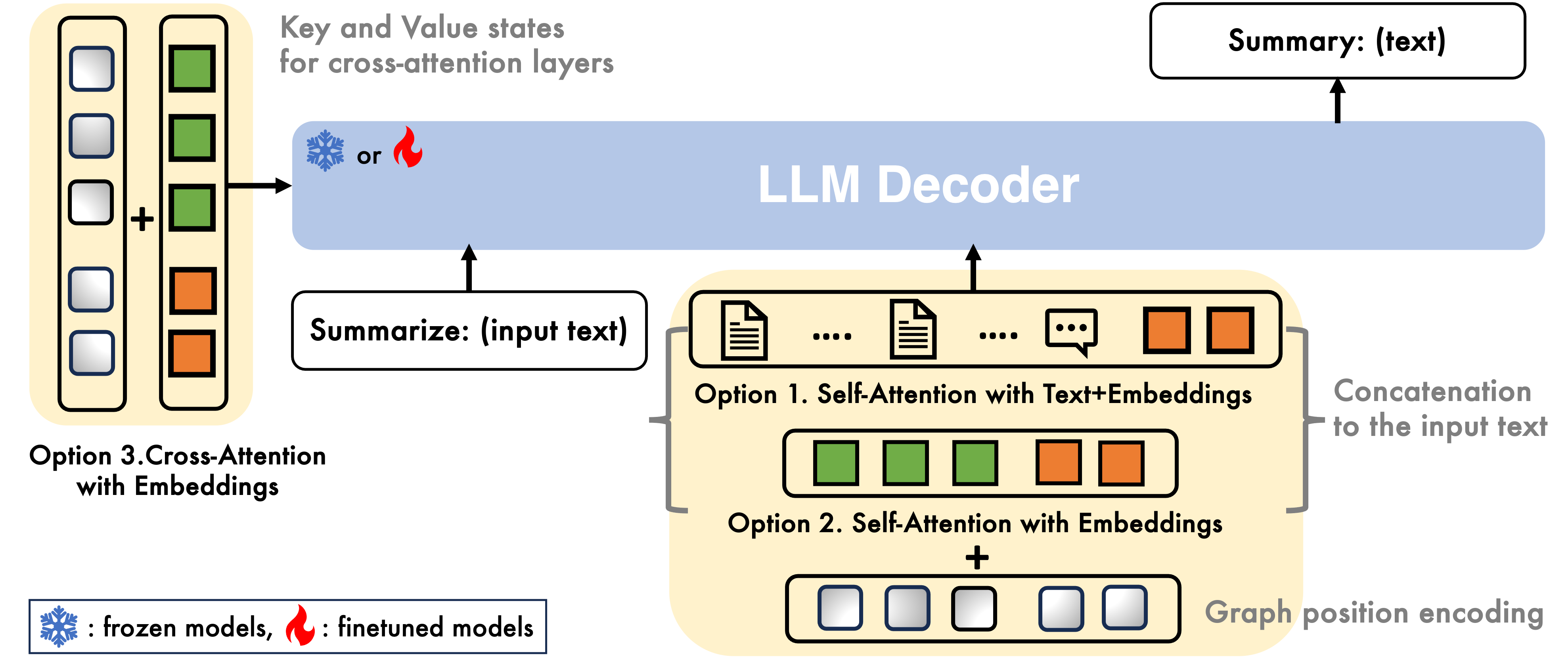}
    }
    \caption
    { \small
    \textbf{Multimodal Graph Learning (MMGL) framework}:
    (a) Multiple multimodal neighbors are given with the input text. 
    (b) Multimodal neighbors are first encoded using frozen vision/text encoders and then aligned to the text-only LM space using $1$-layer MLP mappers. The mappers are trained during LM fine-tuning. Based on the neighbor encoding scheme, texts could be used without any preprocessing (\textit{Self-Attention with Text+Embeddings}) or encoded into embeddings (\textit{Self-Attention with Embeddings} or \textit{Cross-Attention with Embeddings}). Images are always encoded into embeddings to align to the text-only LM space.
    (c) Graph structures among neighbors are encoded as graph position encodings.
    (d) Encoded neighbor information could be infused either by concatenating to the input sequences (\textit{Self-Attention with Text+Embeddings} or \textit{Self-Attention with Embeddings}) or feeding into cross-attention layers (\textit{Cross-Attention with Embeddings}).
    The graph position encodings are added to the input token/text/image embeddings.
    }
    \label{fig:flow}
 \end{figure*}

Given multimodal graphs with text or images on each node, we aim to generate text conditioned on each node and its neighbor nodes.
More specifically, given text input on a target node, pretrained LMs generate free-form text conditioned on the input text and the multimodal context around the target node.
In our multimodal graph learning (MMGL) framework, we first encode each neighbor's information individually using frozen encoders (Figure~\ref{fig:flow:encode}). 
The frozen encoders could be pretrained ViT~\cite{dosovitskiy2020image} or ResNeT~\cite{he2016deep} for images that map pixels to embeddings, and pretrained LMs~\cite{radford2021learning} for texts that map texts to embeddings (similarly for other modalities).
Then, we encode the graph structure around the target node using graph position encodings (Figure~\ref{fig:flow:graph}). 
Finally, the encoded neighbor information with graph position encodings is fed into the pretrained LMs with the input text to generate text conditioned on the multimodal input content (Figure~\ref{fig:flow:decode}).

The framework leaves us with three design spaces:
(1) how can we feed neighbor information to the LMs? 
(2) how can we infuse graph structure information among multimodal neighbors into LMs?
(3) how can we finetune the pretrained LMs to learn from the neighbor context parameter-efficiently? 
In this section, we investigate each problem and discuss possible methodologies we can apply.

\subsection{Research Question 1: Neighbor Encoding}
\label{sec:proposed_work:neighbor_encoding}

Unlike existing multimodal learning, which assumes a single image (corresponding to the input text) as input, multimodal graph learning considers an arbitrary number of neighbor images/texts as input; thus, scalability is the first problem to solve to learn from multiple multimodal neighbors.
In vision-text models, a standard recipe is to first process images with an image encoder (e.g., ViT, ResNet) into image embeddings, then map the embeddings into the text-only LM space, and finally feed them into the LMs. 
Two popular ways to feed image embeddings into LMs are with full self-attention over modalities concatenated across the sequence dimension~\cite{tsimpoukelli2021multimodal} or with cross-modal attention layers~\cite{tsai2019multimodal}.

Motivated by these two approaches, we propose three neighbor encoding methods as follows:
\begin{itemize}[leftmargin=7pt,topsep=0pt,itemsep=-1ex,partopsep=1ex,parsep=1ex]
    \item {
        \textbf{Self-Attention with Text + Embeddings (SA-Text+Embeddings)}: 
        Text neighbors are concatenated as raw texts, while other modalities are first processed by frozen encoders (e.g., ViT for images), and then their embeddings are concatenated to the input sequence. We add a linear mapper that aligns precomputed embeddings into the text space of LLMs.
    }
    \item{
        \textbf{Self-Attention with Embeddings (SA-Embeddings)}: 
        Same as \textit{SA-Text+Embeddings} except text neighbors are also processed by separate frozen encoders, and their embeddings are concatenated to the input sequence. Text encoders could be the same or different from the base LLM model.
    }
    \item{
        \textbf{Cross-Attention with Embeddings (CA-Embeddings)}:
        All neighbors are processed by separate frozen encoders, mapped into the text space by linear mappers, and then fed into cross-attention layers. 
    }
\end{itemize}
In general, when we provide text embeddings instead of raw text, the amount of information the LLMs are able to exploit is bottlenecked by the precomputed embeddings.
However, raw texts introduce scalability issues as the attention mechanism of LMs uses the $O(T^2)$ compute with the sequence length $T$.
Thus, there is a trade-off between computation cost and scalability.
For \textit{SA-Text+Embeddings} and \textit{SA-Embeddings}, we have additional parameters only for mappers that are located outside of the LMs, while \textit{CA-Embeddings} inserts additional cross-attention layers into pretrained LMs and trains them from scratch.
This means \textit{CA-Embeddings} could result in an unstable initial state as the pretrained LLM layers are affected by randomly initialized cross-attention layers.
In Section~\ref{sec:experiments:neighbor_encoding}, we explore these three approaches and discuss their empirical results.

\subsection{Research Question 2: Graph Structure Encoding}
\label{sec:proposed_work:graph}

Given neighbor information, we can simply concatenate neighbor information either as raw texts or embeddings and treat them as a sequence.
But the neighbors have structures among them.
For instance, sections have hierarchical structures, and images are included in certain sections in WikiWeb2M (Figure~\ref{fig:data:nton}).
To encode this graph structure among the neighbor information, we borrow two popular graph position encodings from graph transformers and compare them with sequential position encoding.
\begin{itemize}[leftmargin=7pt,topsep=0pt,itemsep=-1ex,partopsep=1ex,parsep=1ex]
    \item{
        \textbf{Laplacian Position Encoding (LPE)}:
        We exploit Laplacian eigenvectors of neighbors computed from their graph structure as their position encodings. 
    }
    \item{
        \textbf{Graph Neural Networks (GNN)}:
        We first compute neighbor embeddings from frozen encoders and run GNN over the embeddings using the graph structure.
        Then, we use the output GNN embeddings, which encode graph structure information as position encodings. 
    }
\end{itemize}
\textit{LPE} has an additional $1$-layer MLP mapper to map the Laplacian eigenvectors to the text space of LMs.
Parameters used for graph structure encoding (e.g., mappers for \textit{LPE} or \textit{GNN} parameters) are trained with LMs in an end-to-end manner during LM fine-tuning. 
In Section~\ref{sec:experiments:graph}, we explore how well these different position encodings bring additional graph structure information among neighbors into LMs and improve performance. 

\subsection{Research Question 3: Parameter-Efficiency}
\label{sec:proposed_work:peft}
While we need to fine-tune the pretrained LM model for the specific task and newly added neighbor information, full fine-tuning requires high computation costs and also brings inconvenience in sharing MMGL modules when users decide to use neighbor information.
Recently, various parameter-efficient fine-tuning (PEFT) methods have been proposed to fine-tune only a small amount of parameters while preserving the full fine-tuning performance.
We choose three different PEFT models proper for the three neighbor encoding approaches we described above.
\begin{itemize}[leftmargin=7pt,topsep=0pt,itemsep=-1ex,partopsep=1ex,parsep=1ex]
    \item {
        \textbf{Prefix tuning}: 
        When we choose \textit{SA-Text+Embeddings} or \textit{SA-Embeddings} for neighbor encoding, we do not have any newly added parameters but self-attention layers; thus, we can easily apply Prefix tuning~\cite{li2021prefix}, which keeps language model parameters frozen and instead optimizes a sequence of continuous task-specific vectors prepended to the original activation vectors across all layers.
    }
    \item{
        \textbf{LoRA}:
        Like \textit{Prefix tuning}, low-rank adaptation (LoRA)~\cite{hu2021lora} is suitable for \textit{SA-Text+Embeddings} or \textit{SA-Embeddings} neighbor encodings.
        LoRA injects trainable rank decomposition matrices into each layer while freezing the original parameters.
    }
    \item{
        \textbf{Flamingo}:
        For \textit{CA-Embeddings} neighbor encoding, we can directly apply \textit{Flamingo}~\cite{alayrac2022flamingo}, which fine-tunes only newly added cross-attention layers with \textit{tanh} gating to keep the pretrained LM intact at initialization for improved stability and performance.
    }
\end{itemize}
In Section~\ref{sec:experiments:peft}, we explore how well PEFT models preserve the full fine-tuning performance by tuning a small number of parameters.

\section{Experiments}
\label{sec:experiments}
\subsection{WikiWeb2M dataset}
\label{sec:experiments:dataset}

WikiWeb2M dataset~\cite{burns2023wiki} is built for the general study of multimodal content understanding with many-to-many text and image relationships.
Built upon the WIT dataset~\cite{srinivasan2021wit} which contains only image-caption pairs, WikiWeb2M includes the page title, section titles, section text, images and their captions, and indices for each section, their parent section, their children sections, and many more.

In this work, we focus on the section summarization task to generate a single sentence that highlights a particular section’s content. 
The summary is generated given all images and (non-summary) text present in the target and context sections.
We sample $600$k Wikipedia pages randomly from WikiWeb2M for the section summarization task.
In total, the training/validation/test set sizes for the section summarization task are $680$k/$170$k/$170$k, respectively.

\subsection{Experimental Settings}
\label{sec:experiments:settings}

From WikiWeb2M, we can get four types of information for section summarization:
(1) section text, (2) section images, (3) text from page description and other sections, and (4) images from page description and other sections.
We provide information incrementally to LMs to study the effectiveness of multimodal neighbor information: (1) \textit{section text}, 2) \textit{section all} (text + image), 3) \textit{page text} (all text from a Wikipedia page the input section belongs to), and 4) \textit{page all} (all text and images from the Wikipedia page).

We use Open Pre-trained Transformer (OPT-125m)~\cite{zhang2022opt} for the base LM to read the input section text and generate a summary.
For text and image encoders for neighbor information, we use text/image encoders from CLIP~\cite{radford2021learning}.
Following~\cite{raffel2020exploring}, we finetune OPT for $10000$ steps of $125$ batch size with learning rate $10^{-4}$.
The text/image encoders are frozen across all experiments.
We measure BLEU-4~\cite{papineni2002bleu}, ROUGE-L~\cite{lin-2004-rouge}, and CIDEr~\cite{vedantam2015cider} scores on the validation set.
All experiments are run on $4$ Nvidia-RTX 3090 GPUs with $24$GB memory.

\begin{table*}[]
    \caption{
        \small
	\textbf{Effectiveness of neighbor information}:
        As more neighbor information is fed to LMs together with input texts (\textit{section text, section all} => \textit{page text, page all}), generation performance is improved.
        We increase the input sequence length to $1024$ to encode \textit{page text} and \textit{page all} as more information is required to be encoded.
        The best results are colored in red, while the second-best results are colored in blue.
    }
    \label{tab:neighbor}
    \centering
    \small
\begin{tabular}{l|c|ccc}\toprule\midrule
\textbf{Input type} & \textbf{Input length} & \textbf{BLEU-4} & \textbf{ROUGE-L} & \textbf{CIDEr} \\\midrule\midrule
\textbf{Section text} & 512 & 8.31 & 40.85 & 79.68 \\
\textbf{Section all} & 512 & 8.03 & 40.41 & 77.45 \\ \midrule
\textbf{Page text} & 1024 & \cellcolor[HTML]{DAE8FC}9.81 & \cellcolor[HTML]{DAE8FC}42.94 & \cellcolor[HTML]{DAE8FC}92.71 \\
\textbf{Page all} & 1024 & \cellcolor[HTML]{FFCCC9}9.96 & \cellcolor[HTML]{FFCCC9}43.32 & \cellcolor[HTML]{FFCCC9}96.62 \\ \midrule\bottomrule
\end{tabular}
\end{table*}

\subsection{Effectiveness of Neighbor Information}
\label{sec:experiments:neighbor}

We first examine the effectiveness of multimodal neighbor information.
As described in Section~\ref{sec:experiments:settings}, we provide more information incrementally to the base LM: (1) \textit{section text}, (2) \textit{section all} (text + image), 3) \textit{page text}, and 4) \textit{page all} (all texts and images).
Here, we use \textit{Self-Attention with Text+Embeddings (SA-Text+Embeddings)} neighbor encoding across different input types.
For images, we first compute the image embeddings from the frozen CLIP image encoder and concatenate them right after the text of a section each image belongs to preserve the structure.
The results in Table~\ref{tab:neighbor} indicate that \emph{more multimodal neighbor information is helpful:} performance significantly improves when going from \emph{section} to \emph{page} content, and further when adding \emph{page all} content, based on their BLEU-4, ROUGE-L, and CIDEr scores.

\paragraph{Discussion: Missing Modalities.} 
Performance of \textit{section all} decreased slightly from \textit{section text}, despite the addition of section images.
In Wikipedia, not every section has corresponding images.
Thus, in the \textit{section all} case, input to the LMs is inconsistent with some samples having text and images, while other samples only have text.
This points to an important unaddressed \emph{missing modality issue} that is common in the real world, which is not typically encountered in the conventional $1$-to-$1$ multimodal setting, emphasizing the importance of developing MMGL approaches that are robust to the presence of missing modalities.

\begin{table*}[]
    \caption{
        \small
	\textbf{Neighbor encodings in MMGL}:
        We encode multiple multimodal neighbor information using three different neighbor encodings, \textit{Self-Attention with Text+Embeddings} (SA-TE), \textit{Self-Attention with Embeddings} (SA-E), and \textit{Cross-Attenion with Embeddings} (CA-E).
        While SA-TE shows the best performance, SA-TE requires a longer input length ($1024$) to encode texts from neighbors in addition to the original text input, leading to scalability issues.
        The best results are colored in red.
    }
    \label{tab:neighbor_encoding}
    \centering
    \small
\begin{tabular}{l|ccc|ccc|ccc}\toprule\midrule
 & \multicolumn{3}{c}{\textbf{BLEU-4}} & \multicolumn{3}{|c}{\textbf{ROUGE-L}} & \multicolumn{3}{|c}{\textbf{CIDEr}} \\ \midrule
\textbf{Input type} & \textbf{SA-TE} & \textbf{SA-E} & \textbf{CA-E} & \textbf{SA-TE} & \textbf{SA-E} & \textbf{CA-E} & \textbf{SA-TE} & \textbf{SA-E} & \textbf{CA-E} \\ \midrule\midrule
\textbf{Section all} & 8.03 & 7.56 & \cellcolor[HTML]{FFCCC9}8.35 & \cellcolor[HTML]{FFCCC9}40.41 & 39.89 & 39.98 & \cellcolor[HTML]{FFCCC9}77.45 & 74.33 & 75.12 \\
\textbf{Page text} & \cellcolor[HTML]{FFCCC9}9.81 & 8.37 & 8.47 & \cellcolor[HTML]{FFCCC9}42.94 & 40.92 & 41.00 & \cellcolor[HTML]{FFCCC9}92.71 & 80.14 & 80.72 \\
\textbf{Page all} & \cellcolor[HTML]{FFCCC9}9.96 & 8.58 & 8.51 & \cellcolor[HTML]{FFCCC9}43.32 & 41.01 & 41.55 & \cellcolor[HTML]{FFCCC9}96.01 & 82.28 & 80.31 \\ \midrule
\textbf{Max input length} & 1024 & 512 & 512 & 1024 & 512 & 512 & 1024 & 512 & 512 \\ \midrule\bottomrule
\end{tabular}
\end{table*}

\subsection{Neighbor Encoding}
\label{sec:experiments:neighbor_encoding}

We encode multiple multimodal neighbor information using three different neighbor encodings, \textit{Self-Attention with Text+Embeddings} (SA-TE), \textit{Self-Attention with Embeddings} (SA-E), and \textit{Cross-Attenion with Embeddings} (CA-E).
While SA-E and CA-E encode all modalities, including text, into embeddings using frozen encoders, SA-TE encodes text neighbors as they are by concatenating them to the input text sequence.
Thus SA-TE requires longer input sequence lengths ($1024$) to encode additional texts, leading to potential scalability issues.
On the other hand, SA-E and CA-E require one token length to encode one text neighbor, improving scalability with shorter input lengths ($512$).
The results in Table~\ref{tab:neighbor_encoding} indicate that \emph{scalability is traded off with performance:} SA-TE consistently performs better than SA-E and CA-E on different input types at the cost of longer input lengths.

\paragraph{Discussion: Information Loss.}
In conventional multimodal learning with $1$-to-$1$ mappings, SA-TE is commonly used to infuse text input as it is, and image inputs as embeddings are precomputed by frozen encoders~\cite{alayrac2022flamingo,koh2023generating,li2023blip}.
These methods successfully generate texts grounded on the input images, showing image embeddings' effectiveness as input to the pretrained LMs.
However, the performance gap between SA-TE and SA-E in Table~\ref{tab:neighbor_encoding} indicates that text embeddings likely lead to \emph{information loss} in the LMs.
This could be either because the 1-layer MLP mapper that aligns precomputed text embeddings into the text space of the LMs is not expressive enough, or because longer input texts compared to short texts used in the conventional multimodal learning (e.g., one-sentence captions) makes LMs hard to learn from precomputed text embeddings.
From a practical angle, our results illuminate the trade-off between scalability and performance. Meanwhile, our results emphasize the need for more MMGL research to address the challenging issue of information loss when using embeddings to capture text information.

\begin{table*}[]
    \caption{
        \small
	\textbf{Graph structure encoding in MMGL}:
        We encode graph structures among multimodal neighbors using sequential position encodings (\textit{Sequence}), Graph Neural Network embeddings (\textit{GNN}), and Laplacian position encodings (\textit{LPE}). 
        Computed position encodings are added to input token/text/image embeddings and fed into LMs.
        We use \textit{Self-Attention with Embeddings (SA-E)} neighbor encoding and \textit{Prefix tuning} in this experiment.
        The best results are colored in red.
    }
    \label{tab:graph}
    \centering
    \small
\begin{tabular}{l|l|ccc} \toprule\midrule
\textbf{Metric} & \textbf{PEFT} & \textbf{Sequence} & \textbf{GNN} & \textbf{LPE} \\ \midrule\midrule
\multirow{2}{*}{\textbf{BLEU-4}} & \textbf{Prefix tuning} & 6.91 & \cellcolor[HTML]{FFCCC9}6.98 & 6.80 \\
 & \textbf{LoRA} & 7.12 & \cellcolor[HTML]{FFCCC9}7.30 & 7.13 \\ \midrule
\multirow{2}{*}{\textbf{ROUGE-L}} & \textbf{Prefix tuning} & 38.98 & \cellcolor[HTML]{FFCCC9}39.13 & 39.10 \\
 & \textbf{LoRA} & 39.05 & \cellcolor[HTML]{FFCCC9}39.48 & 39.35 \\ \midrule
\multirow{2}{*}{\textbf{CIDEr}} & \textbf{Prefix tuning} & 68.20 & \cellcolor[HTML]{FFCCC9}69.29 & 68.15 \\
 & \textbf{LoRA} & 68.86 & \cellcolor[HTML]{FFCCC9}70.86 & 69.34 \\ \midrule\bottomrule 
\end{tabular}
\end{table*}

\subsection{Graph Structure Encoding}
\label{sec:experiments:graph}

In addition to each modality on neighbors, multimodal graphs contain graph structure information among neighbors.
We encode the graph structures among multimodal neighbors using sequential position encodings (\textit{Sequence}), Graph Neural Network embeddings (\textit{GNN}), and Laplacian position encodings (\textit{LPE}).
Computed position encodings are first mapped to the text space of LMs by $1$-layer MLP, added to input token/text/image embeddings, and fed into LMs.
In Table~\ref{tab:graph}, \textit{GNN} embeddings show the best performance.
Especially, the improvement over \textit{Sequence} position encoding shows the \emph{importance of graph-aware structure encoding methods} in MMGL.

\begin{table*}[]
    \caption{
        \small
	\textbf{Parameter-efficient finetuning in MMGL}:
        We apply \textit{Prefix tuning} and \textit{LoRA} for \textit{Self-Attention with Text+Embeddings (SA-TE)} and \textit{Self-Attention with Embeddings (SA-E)} neighbor encodings.
        For \textit{Cross-Attention with Embeddings (CA-E)} neighbor encoding, we apply \textit{Flamingo}-style finetuning that finetunes only newly added cross-attention layers with gating modules.
        Note that \textit{SA-E} and \textit{CA-E} neighbor encodings have more parameters than \textit{SA-TE} because (frozen) text encoders are added to encode text neighbors.
        The best results are colored in red, while the second-best results are colored in blue.
    }
    \label{tab:peft}
    \centering
    \small
\begin{tabular}{l|l|cc|cc|c}\toprule\midrule
 \multicolumn{2}{l|}{\textbf{Neighbor encoding (max length)}} & \multicolumn{2}{c|}{\textbf{SA-TE (1024)}} & \multicolumn{2}{c|}{\textbf{SA-E (512)}} & \textbf{CA-E (512)} \\ \midrule
\textbf{Metric} & \textbf{Input type} & \textbf{Prefix tuning} & \textbf{LoRA} & \textbf{Prefix tuning} & \textbf{LoRA} & \textbf{Flamingo} \\ \midrule\midrule
\multirow{3}{*}{\textbf{BLEU-4}} 
 & \textbf{Section all} & 6.70 & 6.65 & 6.80 & \cellcolor[HTML]{FFCCC9}7.07 & \cellcolor[HTML]{DAE8FC}6.96 \\
 & \textbf{Page text} & \cellcolor[HTML]{DAE8FC}7.84 & \cellcolor[HTML]{FFCCC9}7.94 & 6.88 & 7.09 & 7.81 \\
 & \textbf{Page all} & \cellcolor[HTML]{FFCCC9}8.21 & \cellcolor[HTML]{DAE8FC}8.18 & 6.91 & 7.12 & 8.12 \\ \midrule
\multirow{3}{*}{\textbf{ROUGE-L}} 
 & \textbf{Section all} & 38.67 & 38.84 & 38.97 & \cellcolor[HTML]{DAE8FC}39.30 & \cellcolor[HTML]{FFCCC9}39.43 \\
 & \textbf{Page text} & \cellcolor[HTML]{DAE8FC}40.61 & \cellcolor[HTML]{FFCCC9}40.98 & 38.38 & 39.69 & 40.29 \\
 & \textbf{Page all} & \cellcolor[HTML]{DAE8FC}41.08 & \cellcolor[HTML]{FFCCC9}41.25 & 38.98 & 39.05 & 40.95 \\ \midrule
\multirow{3}{*}{\textbf{CIDEr}} 
 & \textbf{Section all} & 65.84 & 65.00 & 67.24 & \cellcolor[HTML]{DAE8FC}68.61 & \cellcolor[HTML]{FFCCC9}69.31 \\
 & \textbf{Page text} & \cellcolor[HTML]{DAE8FC}78.12 & \cellcolor[HTML]{FFCCC9}78.60 & 66.55 & 69.26 & 76.20 \\
 & \textbf{Page all} & \cellcolor[HTML]{DAE8FC}81.07 & 80.75 & 68.20 & 68.86 & \cellcolor[HTML]{FFCCC9}82.37 \\ \midrule\midrule
\multicolumn{2}{l|}{\textbf{\# Finetuned parameters}} & 20M & 82M & 20M & 84M & 90M \\ 
\multicolumn{2}{l|}{\textbf{\# Total parameters}} & 230M & 250M & 300M & 320M & 363M \\
\multicolumn{2}{l|}{\textbf{\% Finetuned parameters}} & 9\% & 33\% & 7\% & 26\% & 25\% \\\midrule\bottomrule
\end{tabular}
\end{table*}

\subsection{Parameter-Efficient Fine-Tuning}
\label{sec:experiments:peft}

Full fine-tuning of pretrained LMs requires high computational costs.
For parameter-efficient fine-tuning for MMGL, we study \textit{Prefix tuning} and \textit{LoRA} for \textit{Self-Attention with Text+Embeddings (SA-TE)} and \textit{Self-Attention with Embeddings (SA-E)} neighbor encodings.
For \textit{Cross-Attention with Embeddings (CA-E)} neighbor encoding, we apply \textit{Flamingo}-style finetuning that finetunes only newly added cross-attention layers with gating modules.

The results in Table~\ref{tab:peft} show that \emph{\textit{LoRA} performs better than \textit{Prefix tuning}} for \textit{SA-TE} and \textit{SA-E} neighbor encodings with more fine-tuned parameters ($7-9\%$ for \textit{Prefix tuning} and $26-33\%$ for \textit{LoRA}).
However, \textit{Prefix tuning} still shows comparable performance with \textit{LoRA} using nearly $4$ times fewer parameters with \textit{SA-TE} neighbor encoding.
\textit{Flamingo} with \textit{CA-E} neighbor encoding shows comparable performance with \textit{LoRA} with \textit{SA-TE} neighbor encoding employing the similar numbers of fine-tuned parameters ($82M$ for \textit{LoRA} and $90M$ for \textit{Flamingo}).
Note that \textit{SA-E} and \textit{CA-E} neighbor encodings have more parameters than \textit{SA-TE}, attributed to the inclusion of (frozen) text encoders for text neighbor processing.

In Table~\ref{tab:neighbor_encoding} (without PEFT), it is evident that \textit{CA-E} neighbor encoding lags in performance compared to \textit{SA-TE} neighbor encoding. 
However, when infused with Flamingo, gating modules in Flamingo effectively ensure that the pre-trained LMs remain unaffected by randomly set cross-attention layers at initialization, thereby enhancing the performance of \textit{CA-E}, as shown in Table~\ref{tab:peft} (with PEFT). 
This underscores the pivotal role of strategic initialization when introducing supplementary modules for neighbor encoding in MMGL and when integrating them into the pre-trained LMs.

\section{Conclusion}
\label{sec:conclusion}
In this work, we extend the conventional multimodal learning with $1$-to-$1$ mappings between a pair of modalities into multimodal graph learning (MMGL) with $many$-to-$many$ relations among multiple modalities.
Our MMGL framework is systematically structured around three critical components: (1) neighbor encodings, (2) graph structure encodings, and (3) parameter-efficient fine-tuning.
Through rigorous testing on the WikiWeb2M dataset, we explored different options for each component:  (1) three variations of neighbor encodings, \textit{Self-Attention with Text+Embeddings}, \textit{Self-Attention with Embeddings}, and \textit{Cross-Attention with Embeddings}, highlighting the balance between scalability and performance, (2) three different graph position encodings, \textit{sequence}, \textit{LPE}, and \textit{GNN}, and (3) three PEFT models, \textit{prefix tuning}, \textit{LoRA}, and \textit{Flamingo}, and their trade-off between parameter-efficiency and performance.
Our in-depth analyses and findings aim to lay the groundwork for future MMGL research, igniting further exploration in this field.

\bibliography{myref}
\bibliographystyle{plain}

\end{document}